\documentclass{article} % For LaTeX2e
\usepackage{iclr2027_conference,times}

\usepackage{amsmath,amsfonts,bm}

\def\eqref#1{equation~\ref{#1}}
\def\1{\bm{1}}

\DeclareMathAlphabet{\mathsfit}{\encodingdefault}{\sfdefault}{m}{sl}
\SetMathAlphabet{\mathsfit}{bold}{\encodingdefault}{\sfdefault}{bx}{n}

\usepackage{hyperref}
\usepackage{url}

\newif\ifarxivversion
\arxivversiontrue

\title{AutoSupervision: Closing the Feedback Loop in Scientific Workflows with Grounded Revision Verification}

\author{%
\noindent
\textbf{Haobo Li}\textsuperscript{1},
\textbf{Eunseo Jung}\textsuperscript{3},
\textbf{Wenxiao Zhao}\textsuperscript{2},
\textbf{Feng Liu}\textsuperscript{1},
\textbf{Jiong Wang}\textsuperscript{1},
\textbf{Kaiyi Xu}\textsuperscript{1},\\
\textbf{Zijie Guo}\textsuperscript{1},
\textbf{Zixin Chen}\textsuperscript{3},
\textbf{Ben Fei}\textsuperscript{1},
\textbf{Fenghua Ling}\textsuperscript{1},
\textbf{Lei Bai}\textsuperscript{1}\\
\textsuperscript{1}Shanghai AI Laboratory
\quad
\textsuperscript{2}University of California, Los Angeles\\
\textsuperscript{3}The Hong Kong University of Science and Technology
}

\usepackage{booktabs}
\usepackage{array}
\usepackage{tabularx}
\usepackage{amsmath}
\usepackage{graphicx}

\newcommand{\dataset}{\textsc{AutoSupervision}}

\usepackage{amssymb}

\newcommand{\folderclosed}{\(\scriptscriptstyle\blacktriangleright\)}
\newcommand{\folderopen}{\(\scriptstyle\blacktriangledown\)}

\ifarxivversion
  \iclrfinalcopy
\fi

\begin{document}

\maketitle

\ifarxivversion
  \lhead{Preprint}
\fi

\begin{abstract}
Recent advances in large language models (LLMs) have enabled AI systems to assist scientific research and peer review. However, an essential capability for reliable AI-assisted scientific workflows remains underexplored: verifying whether reviewer feedback leads to meaningful and evidence-supported manuscript improvements. 
We introduce \dataset{}, which evaluates whether scientific manuscript revisions genuinely address reviewer concerns through grounded evidence.
\dataset{} leverages transparent peer-review records as a natural source of supervision, where reviewer comments specify scientific concerns, author responses describe claimed resolutions, and revised manuscripts provide evidence of changes. Given reviewer comments, author responses, and revised manuscripts, models must characterize reviewer concerns, determine whether concerns have been addressed, and identify supporting manuscript evidence. We construct \dataset{} from 56,000 \emph{Nature Communications} articles and corresponding review records. Then we conducted experiments on LLMs, the ablation study, and the case study. Our results show that while LLMs perform well in characterizing reviewer concerns, with GPT-5.5 achieving a score of 0.754, evidence-based verification remains the primary bottleneck, with the best-performing model reaching only 0.501.
\end{abstract}

\section{Introduction}

Scientific discovery is not a one-shot generation process, but an iterative cycle in which claims are proposed, challenged, revised, and re-evaluated in light of evidence and critique~\citep{wang2023scientific}. In scholarly publishing, peer review institutionalizes this cycle: reviewers identify weaknesses, request additional evidence or clarification, authors revise their manuscripts, and reviewers or editors then assess whether the new version has actually addressed the raised concerns~\citep{mulligan2013peer}. Recent advances in LLMs have motivated a vision of AI-assisted scientific discovery, where AI systems participate not only in generating research artifacts but also in evaluating and revising them. This raises a central question for iterative AI scientific workflows: can AI systems verify, with grounded manuscript evidence, whether feedback has actually been resolved in a subsequent revision?

\begin{figure}[t]
    \centering
    \includegraphics[width=0.5\linewidth]{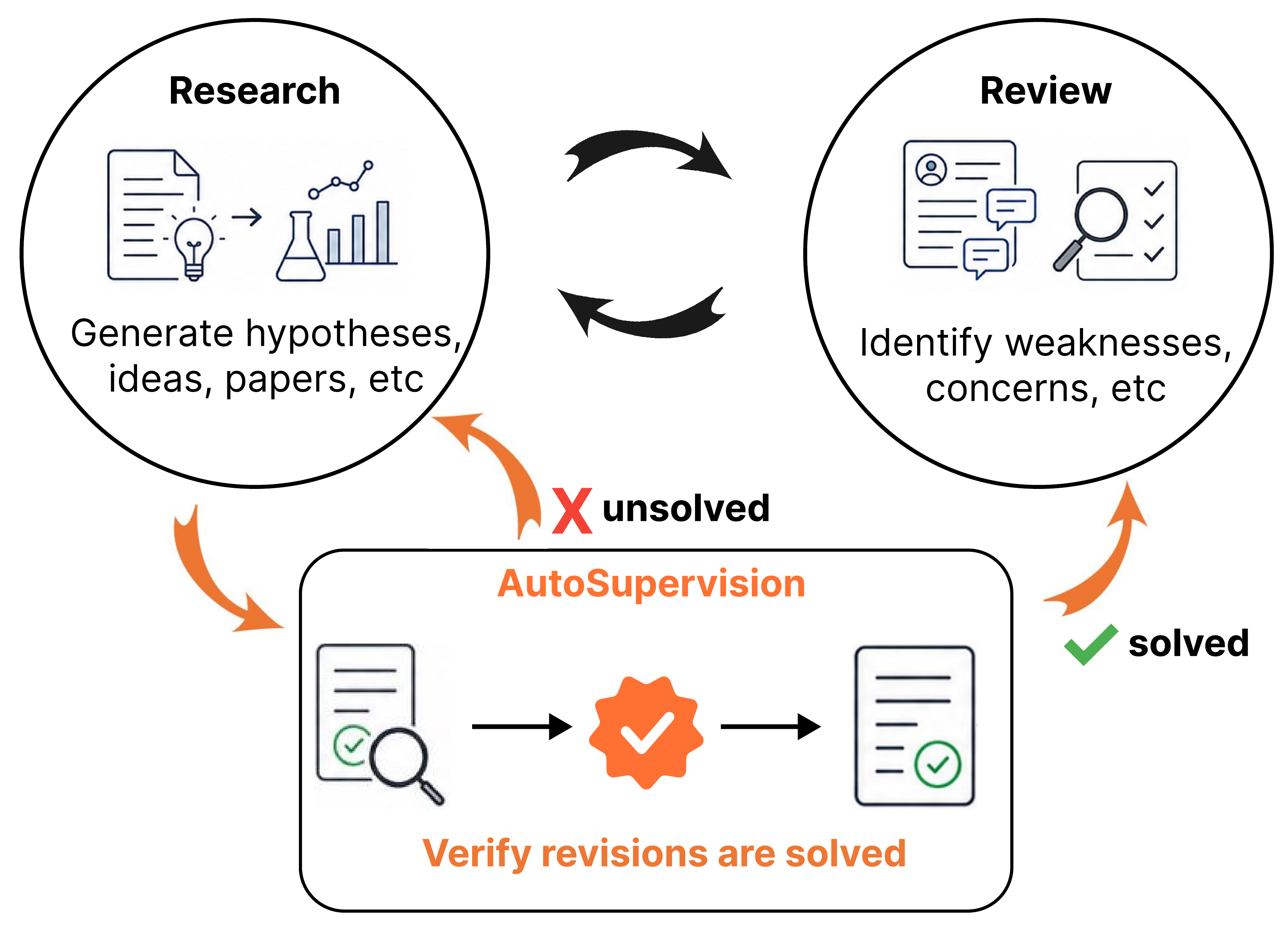}
    \caption{
    \dataset{} for scientific revision verification.
    Black loop: workflow without \dataset{}.
    Orange loop: workflow with \dataset{}: verify whether reviewer concerns are truly addressed.}
    \label{fig:loop}
\end{figure}

% Workflow Gap: AutoResearch and AutoReview cover generation and critique, but not revision verification
Current AI-driven scientific workflows (Figure~\ref{fig:loop}) have begun to address the first two parts of this cycle: generating scientific artifacts and producing feedback on them. AutoResearch systems aim to automate or assist activities such as literature analysis, hypothesis generation, experiment design, coding, and manuscript preparation~\citep{lu2024ai,schmidgall2025agent,ghareeb2026multi}. In parallel, AutoReview systems investigate whether LLM-based agents can evaluate manuscripts, identify weaknesses, generate reviewer comments, and assist human reviewers~\citep{liu2023reviewergpt, zhou2024llmreviewer, latona2024aireviewlottery, cao2025cspaper, jiang2025stanfordagentic, papereview2026}. However, these systems mostly stop at generating artifacts or critiques. They do not directly verify whether a subsequent revision substantively addresses the critique, whether the author's response is reflected in the revised manuscript, or whether the revised text provides sufficient evidence for the paper's scientific claims.

Revision verification poses challenges beyond lexical or semantic matching. Reviewer concerns may target experimental design, claim-evidence alignment, baseline coverage, figure clarity, statistical reporting, or the scope of conclusions, while corresponding revisions may appear as local clarifications, new analyses, additional experiments, revised visualizations, added limitations, or distributed changes across sections. A verifier must therefore recover the intent of the original concern, interpret the author's claimed resolution, and assess whether the revised manuscript provides evidence that substantively addresses the underlying scientific issue. This assessment must also be grounded in the manuscript rather than in the response letter alone. Author responses describe claimed resolutions, but they do not by themselves establish that the manuscript has changed in the relevant way. A response may overstate the revision, address only part of the concern, or describe a change whose evidential support remains insufficient. Reliable verification must therefore jointly reason over the reviewer concern, the author response, and concrete manuscript evidence.

% Benchmark gap: existing peer-review resources do not define concern-level, evidence-grounded revision verification
Existing peer-review datasets provide valuable resources for modeling review artifacts, but they do not directly support grounded revision verification. PeerRead, NLPeer, and MOPRD include combinations of papers, reviews, rebuttals, manuscript versions, meta-reviews, and decisions~\citep{kang2018peerread, dycke2023nlpeer, lin2023moprd}. Related resources study meta-review generation, review discussion modeling, review-rebuttal argument relations, review quality assessment, and substantiation of review statements~\citep{cheng2020ape, shen2022mred, kennard2022disapere, ghosal2022peerreviewanalyze, guo2023substantiation}. These datasets have enabled substantial progress in computational peer-review research, but they are primarily organized around reviews, rebuttals, decisions, or discussion structures. They do not define concern-level revision verification as a benchmark task, where a model must jointly read the reviewer concern, the author response, and the revised manuscript to decide whether the concern was resolved and to identify the manuscript evidence supporting that judgment. This leaves the following task largely unaddressed:
\begin{quote}
Given a reviewer's concern, an author's response, and a revised manuscript, can a model characterize the concerns and whether the concerns have been addressed and identify the exact manuscript evidence supporting this judgment?
\end{quote}

% Our work, \dataset{}, addresses this problem by introducing a benchmark for grounded verification of scientific manuscript revisions. 
% The benchmark is defined at the level of individual reviewer concerns within a revision episode and evaluates how fully each concern was addressed using the author response and the revised manuscript. This concern-level formulation measures revision verification separately from the task of identifying concerns from reviewer comments, thereby ensuring that model performance reflects the ability to assess revisions rather than errors arising from the identification of reviewer comments.

% The benchmark is constructed from transparent peer-review records of \emph{Nature Communications}, which provides peer-review files containing reviewer comments and author responses alongside published articles~\citep{naturecommunications2022transparent}. These records enable studying how scientific feedback is transformed into manuscript revisions across diverse research domains. We collect 56,000 article-level records and further extract revision episodes and concern-level verification instances for benchmark evaluation.
% We evaluate a broad range of LLMs, the supervised model, and the agentic system on \dataset{}. Our results show that modern models achieve strong characterization ability, but still struggle to verify whether revisions genuinely address reviewer concerns and to identify supporting grounding evidence.

% Present study: AutoSupervision benchmark construction and evaluation
To make this question benchmarkable, we introduce \dataset{}, a benchmark for AutoSupervision that evaluates concern-level, evidence-grounded verification of scientific manuscript revisions. Each instance is centered on a reviewer concern within a revision episode. Given the concern, the corresponding author response, and the revised manuscript, a model must characterize the concern, determine whether it has been resolved, and identify the manuscript evidence supporting its judgment. This formulation isolates revision assessment from concern discovery, so model performance reflects the ability to verify revisions rather than the ability to extract reviewer comments.

We construct \dataset{} from transparent peer-review records of \emph{Nature Communications}, which provide reviewer comments and author responses alongside published articles~\citep{naturecommunications2022transparent}. From 56,000 article-level records, we extract revision episodes and concern-level verification instances by aligning reviewer concerns, author responses, and manuscript evidence. We then evaluate LLMs, supervised models, and agentic verification systems. The results show that modern models can often characterize reviewer concerns, but still struggle to determine whether revisions genuinely resolve them and to ground their judgments in manuscript evidence.

Our contributions are summarized as follows:

\begin{itemize}
    \item We define AutoSupervision as a new evaluation problem for AI-assisted scientific workflows and introduce \dataset{}, a benchmark for concern-level, evidence-grounded verification of scientific manuscript revisions.

    \item We construct a large-scale dataset from 56,000 transparent peer-review article records from \emph{Nature Communications} and develop a systematic pipeline to align reviewer concerns, author responses, and manuscript revisions, yielding 8,790 episode-level instances with resolution labels and evidence annotations.

    \item We provide comprehensive evaluations across LLMs, supervised models, agentic systems, and ablation studies, revealing the remaining challenges in evidence-grounded revision assessment.

\end{itemize}

\section{Related Work}
\paragraph{Resources for computational peer review.}
Computational studies of peer review have produced several valuable resources for modeling scientific evaluation. PeerRead pairs paper drafts with reviews and acceptance decisions, enabling research on review generation, review analysis, and acceptance prediction~\citep{kang2018peerread}. NLPeer unifies peer-review datasets with structured representations, licensing information, and reviewing-assistance tasks~\citep{dycke2023nlpeer}. MOPRD introduces a multidisciplinary open peer-review dataset containing review comments, rebuttal letters, manuscript versions, meta-reviews, and decisions~\citep{lin2023moprd}. Peer Review Analyze further annotates ICLR reviews with paper-section correspondence, aspect categories, statement purposes, and significance labels ~\citep{ghosal2022peerreviewanalyze}. These datasets enable computational studies of peer review, but primarily focus on review artifacts, paper-level decisions, or properties of reviewer feedback rather than whether feedback is successfully incorporated into revised manuscripts.

\paragraph{Modeling scientific feedback and revision processes.}
Prior work has studied how reviewers and authors interact through review comments and response documents. APE introduced argument pair extraction between peer reviews and rebuttals, enabling analysis of reviewer arguments and author responses~\citep{cheng2020ape}. DISAPERE annotated discourse structures in review-rebuttal pairs and characterized author stances toward reviewer arguments~\citep{kennard2022disapere}. MReD studied structured meta-review generation by modeling review discussions and decision rationales~\citep{shen2022mred}. These efforts provide important foundations for understanding scientific feedback exchange. However, they mainly treat rebuttals as the endpoint of the interaction. In contrast, \dataset{} incorporates the revised manuscript as an additional component and evaluates the complete feedback chain from reviewer concern to author response, to evidence of manuscript change.

\begin{figure*}[t]
\centering
\includegraphics[width=\textwidth]{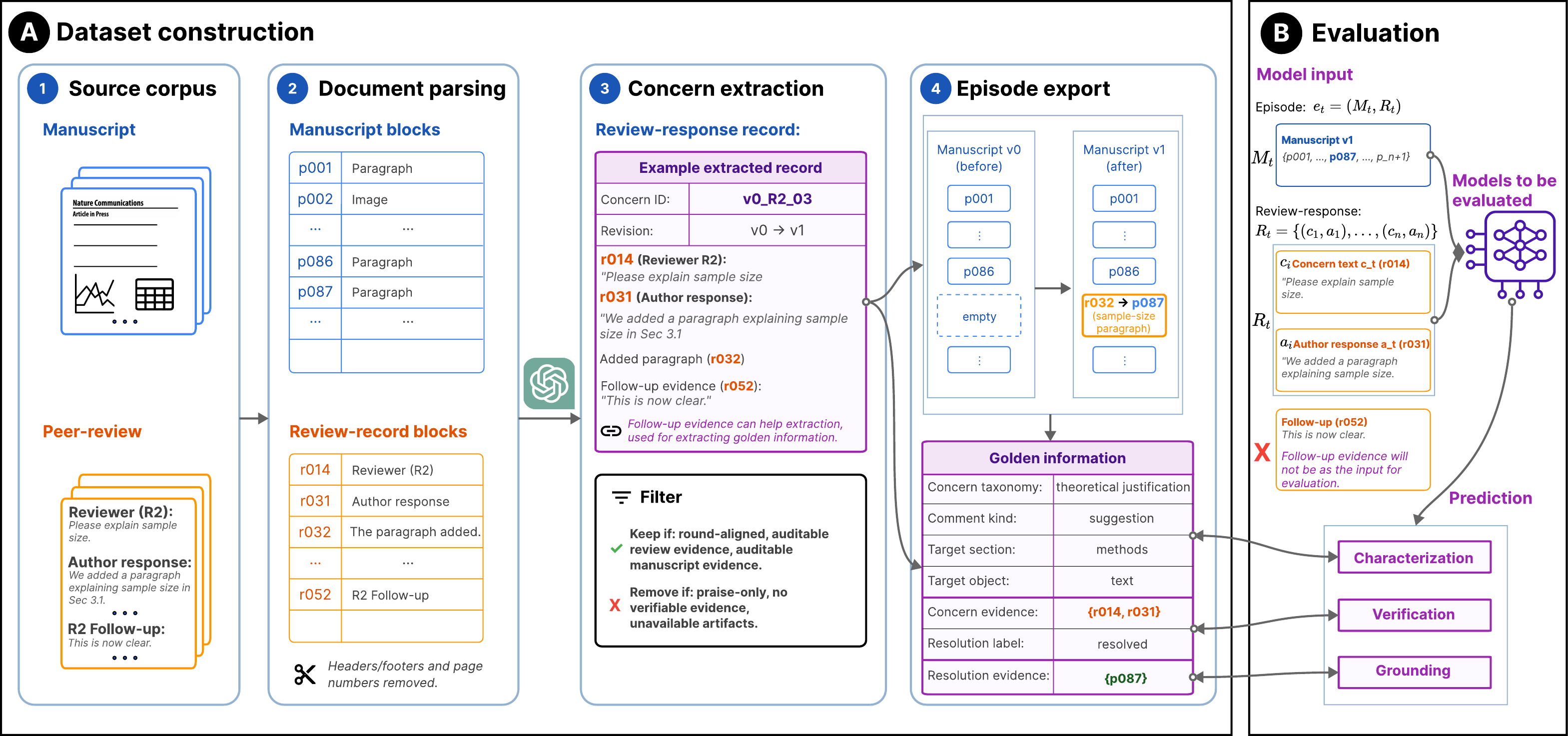}
\caption{\dataset{} construction and evaluation pipeline. First, we pair articles and peer reviews (A1), parse them into blocks with identifiers (A2), extract concerns (A3), and export revision episodes for concern characterization, resolution verification, and evidence grounding (A4). After building \dataset{}, we benchmark existing LLMs on this task (B).}
\label{fig:pipeline}
\end{figure*}

\paragraph{LLMs for scientific research and review assistance.}
Recent advances in LLMs have motivated AutoResearch systems that automate or assist scientific research such as literature analysis, hypothesis generation, experimentation, coding, and manuscript preparation~\citep{lu2024ai,schmidgall2025agent,ghareeb2026multi}. In parallel, AutoReview systems explore the use of LLMs to evaluate manuscripts and provide scientific feedback. ReviewerGPT investigates LLM-based generation of reviewer comments and manuscript critiques~\citep{liu2023reviewergpt}. Other studies evaluate the quality and usefulness of LLM-generated reviews and research-paper feedback~\citep{zhou2024llmreviewer,liang2024can}. CSPaper Review provides fast, rubric-faithful feedback aligned with conference review criteria~\citep{cao2025cspaper}. PapeReview offers an agentic pre-submission review workflow that produces structured feedback on manuscript clarity, methodology, and contribution~\citep{papereview2026}. The Stanford Agentic Reviewer uses an agentic workflow that retrieves relevant work from arXiv and synthesizes it with manuscript content to generate actionable feedback~\citep{jiang2025stanfordagentic}. REVAS has also been deployed in ACL Rolling Review to provide reviewers with feedback on the quality of their reviews~\citep{arr2026revas}.

These systems automate or assist the generation and improvement of scientific feedback. However, they do not evaluate whether reviewer feedback is subsequently incorporated into a revised manuscript. \dataset{} addresses this complementary stage by characterizing reviewer concerns, verifying their resolution, and grounding the resulting judgments in manuscript evidence.

% Table~\ref{tab:related-comparison} summarizes the distinction. Prior resources are valuable for studying review text, reviewing quality, decisions, and rebuttal discourse. \dataset{} instead treats revision as the object of evaluation: the model must read the reviewer, the author, and the revised scientific manuscript together.

% \begin{table*}[t]
% \centering
% \setlength{\tabcolsep}{3pt}
% \begin{tabularx}{\textwidth}{@{}lXXX@{}}
% \toprule
% Resource 
% & Feedback Modeling
% & Revision Verification
% & Evidence Grounding \\
% \midrule
% PeerRead \citep{kang2018peerread}
% & Review artifacts
% & No
% & No \\

% NLPeer \citep{dycke2023nlpeer}
% & Review resources
% & No
% & No \\

% MOPRD \citep{lin2023moprd}
% & Reviews, rebuttals, versions
% & No explicit task
% & No \\

% APE / DISAPERE 
% \citep{cheng2020ape,kennard2022disapere}
% & Review-response interaction
% & No
% & No \\

% LLM Review Systems
% \citep{liu2023reviewergpt,zhou2024llmreviewer,liang2024can}
% & Review generation and assistance
% & No
% & No \\

% \dataset{}
% & Reviewer concern and response
% & Yes
% & Yes \\

% \bottomrule
% \end{tabularx}
% }
% \caption{Comparison of capabilities studied by existing peer-review resources and \dataset{}.}
% \label{tab:related-comparison}
% \end{table*}

\section{Task Definition}

\subsection{Scientific Revision Loop}

We formalize scientific revision as an interaction among three roles: an author process, a reviewer process, and an AutoSupervision model. Let \(\mathcal{A}\) denote the author, \(\mathcal{R}\) denote the reviewer process, and \(\mathcal{S}_{\theta}\) denote the AutoSupervision model to be evaluated. At revision round \(t\), the manuscript before revision is \(M_{t-1}\). A reviewer observes the manuscript and produces a set of reviewer concerns:
\begin{equation}
    C_t = \mathcal{R}(M_{t-1}) = \{c_{t,1}, \ldots, c_{t,n_t}\},
\end{equation}
where \(n\) is the number of concerns at round \(t\).
The author then responds to these concerns and revises the manuscript:
\begin{equation}
    (M_t, A_t) = \mathcal{A}(M_{t-1}, C_t),
\end{equation}
where \(M_t\) is the post-revision manuscript and
\(A_t=\{a_{t,1}, \ldots, a_{t,n_t}\}\) denotes the author responses or rebuttals corresponding to the reviewer concerns.

For each concern \(c_{t,i}\), we define a review-response record
\begin{equation}
    r_{t,i}=(c_{t,i}, a_{t,i}),
\end{equation}
and let \(R_t=\{r_{t,1}, \ldots, r_{t,n_t}\}\) be the set of records associated with revision round \(t\). In the benchmark, \(\mathcal{A}\) and \(\mathcal{R}\) are not models to be trained. They are observed through public transparent peer-review records: reviewer comments instantiate \(C_t\), author responses instantiate \(A_t\), and published manuscript revisions instantiate \(M_t\). The task is to evaluate whether \(\mathcal{S}_{\theta}\) can audit the result of this interaction.

\subsection{Round-Level Episodes}

Peer-review data contains multiple manuscript versions across revision rounds. We therefore define a \emph{revision episode} as a transition between two
consecutive manuscript versions, such as \texttt{v0\_to\_v1} or \texttt{v1\_to\_v2}. Formally, an episode is
\begin{equation}
    e_t=(M_t, R_t).
\end{equation}
This formulation evaluates each concern together with the author response and manuscript state available at the corresponding revision stage, rather than aggregating information across the entire review history.

\subsection{AutoSupervision Prediction Task}

Given an episode \(e_t\), the AutoSupervision model reads the post-revision manuscript together with the full set of review-response records from that round and predicts a set of structured outputs \(\hat{Y}_t\):
\begin{equation}
    \hat{Y}_t
= \mathcal{S}_{\theta}(e_t)
= \{\hat{y}_{t,i}\}_{i=1}^{n_t}.
\end{equation}
Each predicted element corresponds to one record
\(r_{t,i}\). It contains concern-characterization fields, revision-verification fields, review-record evidence block IDs, and revised-manuscript grounding block IDs.

The gold output for the same episode \(Y^{*}_t\) is
\begin{equation}
    Y^{*}_t=\{y^{*}_{t,i}\}_{i=1}^{n_t}.
\end{equation}
Evaluation compares \(\hat{Y}_t\) with \(Y^{*}_t\) for every episode, aligning predictions by the fixed review-response record identifiers.

This task is different from automatic review generation and automatic manuscript revision. The model is not asked to generate reviewer concerns \(C_t\), write author responses \(A_t\), or produce a revised manuscript \(M_t\). Instead, \(\mathcal{S}_{\theta}\) audits the closed feedback loop: given what the reviewer requested, what the author claimed, and what appears in the revised manuscript, it must determine whether the concern was addressed and cite the exact evidence supporting that judgment.

The full prediction schema and label definitions are provided in Appendix~\ref{sec:prediction-evaluation-details}.

\section{Dataset Construction}
\label{sec:benchmark-construction}

\subsection{Pipeline}

Figure~\ref{fig:pipeline} (A) summarizes the dataset construction pipeline.

\paragraph{Source Corpus and Splits} 
\label{corpus}
We collect 56,000 published articles and corresponding transparent peer-review files from \emph{Nature Communications}. Within each publication year, papers are assigned to training, validation, and test partitions using an approximately $8{:}1{:}1$ ratio. All review rounds and manuscript states from the same paper remain in the same partition, preventing cross-round information leakage while preserving the publication-year distribution across splits.

\begin{table}[t] 
\centering 
\begin{tabular}{lr} 
\toprule 
Statistic & Count \\ 
\midrule 
Source-corpus papers & 56,000 \\ 
\hspace{12pt} \folderclosed\ Training split & 44,796 \\ 
\hspace{12pt} \folderclosed\ Validation split & 5,594 \\ 
\hspace{12pt} \folderopen\ Test split & 5,610 \\ 
\hspace{24pt} \folderopen\ Sampled test papers & 557 \\ 
\hspace{40pt} Revision episodes & 847 \\ 
\hspace{40pt} Unique reviewer concerns & 6,543 \\ 
\hspace{40pt} Episode-level concern instances & 8,790 \\ 
\bottomrule 
\end{tabular} 
\caption{Source-corpus and benchmark statistics.} 
\label{tab:dataset-statistics} 
\end{table} 

\paragraph{Document Parsing} 
Both the published articles and their corresponding peer-review files are downloaded in PDF format. Because PDFs primarily encode visual layout rather than structured document content, we first parse them into ordered, typed blocks using MinerU \citep{wang2024mineru}. The resulting blocks include paragraphs, section headings, figures, tables, equations, references, captions, and other scientific content. Each manuscript block receives a stable identifier with the prefix \texttt{p}, while each peer-review block receives an identifier with the prefix \texttt{r}. 
% These identifiers are preserved throughout annotation, filtering, episode construction, model inference, and evaluation. During parsing, we remove layout artifacts such as page headers, footers, page numbers, and post-acceptance metadata while retaining the scientific content and its document order. Stable block identifiers allow annotations and model predictions to cite auditable evidence units rather than free-form text spans or approximate document locations.

\paragraph{Concern Extraction} 
For each sampled paper, we use GPT-5.5~\citep{openai2026gpt55} to process the published article and its peer review. A peer review may contain multiple rounds of reviewer comments, author responses, and reviewer follow-up messages. The model extracts candidate concern threads, links each reviewer concern to its corresponding author response and follow-up discussion, and assigns the resulting review-response record to the revision episode in which the response was made.

During benchmark construction, the model has access to the complete public review history. This retrospective context provides useful evidence. For example, a later reviewer follow-up may explicitly confirm that a concern was addressed, restate an unresolved issue, or narrow the remaining objection after examining the authors' revision. Reviewer follow-up information is used as gold information only during benchmark construction for concern-response alignment and quality filtering. It is not provided as input during evaluation.

Each extracted record preserves the original reviewer concern, the corresponding author response, the normalized reviewer identifier, the assigned revision round, and the supporting review-block identifiers. It also includes the concern-characterization, resolution, and target-scope annotations defined by the benchmark schema.

\paragraph{Episode Export} 

We apply conservative filtering to ensure that retained instances represent actionable and auditable scientific revision requests. We exclude: 
\begin{itemize} 
    \item praise-only comments and broad paper-level assessments without a specific revision request; 
    \item concerns without verifiable reviewer-comment evidence;
    \item concerns whose targets exist only in the rebuttal letter, an external repository, or other unavailable artifacts.
\end{itemize} 

When a concern recurs across multiple revision rounds, we retain the reviewer comment and corresponding author response for each relevant episode while storing later follow-up discussion. This prevents short follow-up statements, such as ``OK'' or ``Addressed,'' from replacing the original scientific concern and response context. 

For each valid transition from manuscript version $t-1$ to version $t$, we export one revision episode \( e_t \). 
Each episode includes the complete set of manuscript blocks available after the revision and only the review information associated with the corresponding round. Information from later rounds may support annotation and filtering but is never exposed as input for an earlier evaluation episode.

\subsection{Dataset Statistics} 
\label{sec:dataset-statistics} 
% The source corpus contains 56,000 paired article-review records, including 44,796 training papers, 5,594 validation papers, and 5,610 test papers. 

For the benchmark evaluation in this paper, we sample 10\% of the test papers from each year using a fixed random seed, yielding 557 candidate papers. After review-response extraction, revision alignment, and quality filtering, the resulting benchmark contains 847 revision episodes. Across these episodes, \dataset{} contains 6,543 unique point-level reviewer concerns. Because a concern may remain active or recur across multiple revision rounds, the unique concerns correspond to 8,790 episode-level concern instances. These episode-level instances are the primary units used for benchmark evaluation. 
More details about dataset distribution are provided in Appendix~\ref{sec:distribution}.
To evaluate the robustness of our annotations to the choice of model, we measure inter-model agreement among the annotation outputs, as reported in Appendix~\ref{sec:agreement}.

\begin{table*}[t]
\centering
\small
\setlength{\tabcolsep}{3pt}
\begin{tabular}{lrrrrr}
    \toprule
    System & Coverage & Characterization & Verification & Grounding & Overall \\
    \midrule
    Claude-Opus-4.8~\citep{anthropic2026claudeopus48} & 0.977 & 0.691 & \textbf{0.501} & \textbf{0.719} & \textbf{0.637} \\
    GPT-5.5~\citep{openai2026gpt55} & 1.000 & \textbf{0.754} & 0.455 & 0.702 & \textbf{0.637} \\
    GLM-5.2~\citep{zai2026glm52} & 1.000 & 0.698 & 0.475 & \underline{0.711} & \underline{0.628} \\
    Gemini-3.1-Pro~\citep{googledeepmind2026gemini31pro} & 0.997 & 0.698 & 0.473 & 0.700 & 0.624 \\
    Qwen3.7-max~\citep{qwen2026qwen37} & 1.000 & 0.701 & 0.463 & 0.699 & 0.621 \\
    SFT Qwen3.5-9B~\citep{qwen2026qwen359b} & 1.000 & \underline{0.708} & \underline{0.490} & 0.643 & 0.614 \\
    DeepSeek-V4-Pro~\citep{xu2026deepseek} & 1.000 & 0.668 & 0.451 & 0.575 & 0.565 \\
    Kimi-K2.6~\citep{moonshot2026kimik26} & 1.000 & 0.653 & 0.455 & 0.571 & 0.560 \\
    DeepSeek-V3.2~\citep{liu2025deepseek} & 1.000 & 0.677 & 0.396 & 0.551 & 0.541 \\
    MiniMax-2.7~\citep{minimax2026m2} & 1.000 & 0.626 & 0.373 & 0.563 & 0.521 \\
    DeepSeek-V4-Flash~\citep{xu2026deepseek} & 1.000 & 0.629 & 0.399 & 0.443 & 0.490 \\
    Intern-S2-35B~\citep{internlm2026interns2} & 1.000 & 0.594 & 0.351 & 0.470 & 0.472 \\
    Qwen3.5-9B~\citep{qwen2026qwen359b} & 1.000 & 0.563 & 0.376 & 0.451 & 0.463 \\
    GPT-4o-mini agentic~\citep{openai2024gpt4omini} & 0.998 & 0.573 & 0.300 & 0.336 & 0.403 \\
    GPT-4o-mini~\citep{openai2024gpt4omini} & 0.933 & 0.497 & 0.216 & 0.085 & 0.266 \\
    \bottomrule
\end{tabular}
\caption{Main \dataset{} performance results. Coverage is concern-level prediction coverage, and Overall is the equal-weight average of characterization, verification, and manuscript grounding. Bold and underlined values indicate the best and second-best performance per backbone model, respectively.}
\label{tab:main-results}
\end{table*}

% \subsection{Concern and Revision Distributions} 
% Figure~\ref{fig:concern-distributions} summarizes the composition of \dataset{} across reviewer concern taxonomy, comment kind, target section, and target object. 
% The benchmark includes heterogeneous concerns involving writing, claim-evidence support, theoretical justification, etc.
% Concerns also target different manuscript sections and require different forms of revision evidence, ranging from localized textual clarifications to new analyses and experimental results. These distributions reflect the central difficulty of revision verification: adequate resolution cannot be determined using a single form of lexical or structural matching. Models must interpret the reviewer request, relate it to the author's response, and locate the corresponding evidence in the revised manuscript.

\section{Evaluation}
\subsection{Evaluation Pipeline}

Figure~\ref{fig:pipeline} (B) summarizes the Evaluation pipeline. We evaluate models under a unified structured prediction setting. All models must output JSON conforming to the benchmark schema. Missing concern IDs are scored as missing predictions, and exact concern-set match is reported separately from label and grounding scores. The evaluator aligns predictions by the fixed input concern IDs, so systems do not receive credit for identifying concerns outside the provided evaluation set.

The benchmark evaluates three complementary capabilities: Characterization, Verification, and Grounding. The Characterization score averages comment-kind macro-F1, point-type macro-F1, and a hierarchical target-scope score over target domain, section, and object. The Verification score averages resolution-label macro-F1, paper-status macro-F1, and micro-F1 over evidence block IDs supporting the resolution judgment. The Grounding score is computed as micro-F1 over target paper block IDs, which identify the revised-manuscript evidence blocks where a reviewer concern is addressed. 
Please refer to Appendix \ref{sec:prediction-evaluation-details} for details. 
The overall score assigns equal weights to the three components:

\[
S_{\mathrm{overall}}
=
\frac{1}{3}
\left(
S_{\mathrm{char}}
+
S_{\mathrm{ver}}
+
S_{\mathrm{ground}}
\right).
\]

% The baseline contains three groups. First, we evaluate full-context single-shot LLMs that receive the complete input context. Second, we conduct GPT-5.5 ablations including concern-only, response-only, manuscript-only, and a retrieve-then-read setting using BM25 retrieval over manuscript blocks followed by LLM inference. Third, we evaluate an agentic decomposition baseline based on GPT-4o-mini, which separates characterization, retrieval, grounding, verification, and consistency checking into individual stages.

\subsection{Evaluated Models}
We evaluate \dataset{} on a diverse set of recent LLMs, covering both closed-source and open-source models. The evaluated models span different model families and scales, enabling a comprehensive assessment of current capabilities for reviewer concern analysis.

\paragraph{Closed-source models.}
We include several proprietary frontier models, including Claude-Opus-4.8~\citep{anthropic2026claudeopus48}, GPT-5.5~\citep{openai2026gpt55}, and Gemini-3.1-Pro~\citep{googledeepmind2026gemini31pro}. These models represent state-of-the-art general-purpose language models with strong reasoning and instruction-following capabilities. We additionally evaluate GPT-4o-mini~\citep{openai2024gpt4omini} and its agentic variant to examine the performance gap between earlier-generation compact models and recent frontier systems.

\paragraph{Open-source models.}
We evaluate a broad range of open-source models, including GLM-5.2~\citep{zai2026glm52}, Qwen3.7-max~\citep{qwen2026qwen37}, DeepSeek-V4-Pro~\citep{xu2026deepseek}, Kimi-K2.6~\citep{moonshot2026kimik26}, DeepSeek-V3.2~\citep{liu2025deepseek}, MiniMax-2.7~\citep{minimax2026m2}, DeepSeek-V4-Flash~\citep{xu2026deepseek}, Intern-S2-35B~\citep{internlm2026interns2}, and Qwen3.5-9B~\citep{qwen2026qwen359b}. We also include an instruction-tuned Qwen3.5-9B model to investigate the impact of supervised fine-tuning on reviewer concern understanding.

\subsection{Main Results}
\begin{figure}[t]
\centering
\includegraphics[width=0.5\columnwidth]{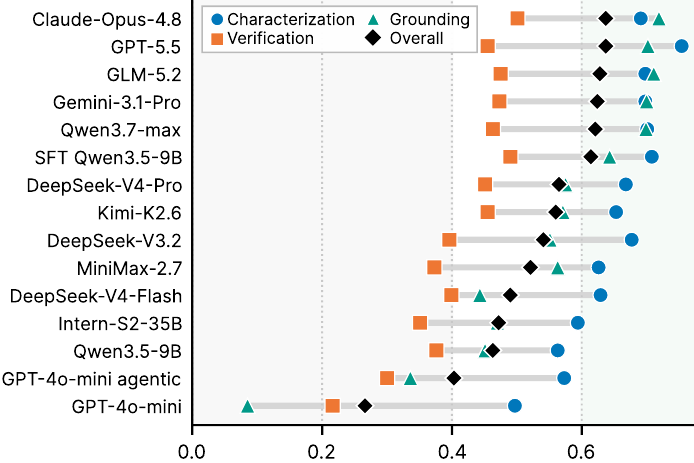}
\caption{Model performance across characterization, verification, and grounding. Gray segments show the three component scores, and the black diamond denotes the equal-weight overall score.}
\label{fig:main-metric-profile}
\end{figure}

Table~\ref{tab:main-results} and Figure~\ref{fig:main-metric-profile} summarize benchmark performance across closed-source and open-source models. Overall, the results reveal a substantial capability gap between identifying reviewer concerns and performing evidence-based assessment of revisions. While modern models can almost always extract and characterize reviewer concerns, accurately verifying whether a revision resolves the concern remains a significant challenge.

Among all evaluated systems, Claude-Opus-4.8 and GPT-5.5 achieve the highest overall performance (both 0.637), followed by GLM-5.2 (0.628) and Gemini-3.1-Pro (0.624). Interestingly, the performance difference between the strongest closed-source and open-source models is relatively small, with GLM-5.2 approaching frontier proprietary systems. However, their remaining limitations are primarily concentrated in deeper reasoning tasks that require evaluating the relationship between a claimed revision and the underlying manuscript evidence.

\paragraph{Coverage}
Coverage is concern-level prediction coverage. Nearly all models achieve perfect or near-perfect coverage, indicating that modern LLMs rarely fail to identify the presence of reviewer concerns. The only notable degradation occurs for smaller or earlier-generation models, such as GPT-4o-mini. This suggests that concern extraction is largely a solved capability and is no longer the primary bottleneck of reviewer-assistance systems.

\paragraph{Characterization}
Characterization measures whether a model correctly identifies the type and scope of a reviewer concern. GPT-5.5 achieves the strongest performance (0.754), while most frontier models exceed 0.65. The high characterization scores, combined with near-perfect coverage, indicate that models possess strong semantic understanding of reviewer feedback.

\paragraph{Verification}
Verification is consistently the most challenging dimension. Claude-Opus-4.8 obtains the highest verification score (0.501), followed by the Qwen3.5-9B SFT model (0.490) and GLM-5.2 (0.475). This pattern is notable because verification does not simply require language understanding; it requires multi-step reasoning over the reviewer concern, the claimed revision, and the manuscript content. The results suggest that current models often recognize the intent of a revision but struggle to determine whether the provided evidence is sufficient, complete, and scientifically relevant.

\paragraph{Grounding}
Manuscript grounding evaluates whether models can identify supporting evidence. Claude-Opus-4.8 achieves the highest grounding score (0.719), followed closely by GPT-5.5 (0.702), GLM-5.2 (0.711), and Gemini-3.1-Pro (0.700). In contrast, smaller models show substantial degradation, with GPT-4o-mini achieving only 0.085. The large performance gap highlights that grounding requires more than generating plausible explanations: models must accurately retrieve and connect manuscript-level evidence to reviewer concerns.

Overall, these results demonstrate that \dataset{} evaluates capabilities beyond reviewer feedback understanding. Existing LLMs already achieve strong performance in detecting and characterizing concerns, but they remain limited in evidence-based verification and grounding.

\begin{table}[t]
\centering
\setlength{\tabcolsep}{1mm} 
\begin{tabular}{lrrrr}
\toprule
Setting & Char. & Verif. & Ground. & Overall\\
\midrule
Concern-only & 0.676 & 0.042 & 0.000 &  0.239 \\
Concern+Response & 0.696 & 0.236 & 0.000 & 0.311 \\
Concern+Manuscript & 0.719 & 0.394 & 0.633 & 0.582 \\
Full context & 0.754 & 0.455 & 0.702 & 0.637 \\
\bottomrule
\end{tabular}
\caption{Ablation study. GPT-5.5 performance under different input configurations.}
\label{tab:ablation}
\end{table}

\subsection{Impact of Agentic Inference}
Beyond scaling model size, we evaluate whether a fixed structured inference pipeline can improve performance. The agentic baseline uses GPT-4o-mini in a three-stage pipeline for concern characterization, manuscript grounding, and resolution verification. Between characterization and grounding, we add a deterministic manuscript-block retriever that narrows the full manuscript to a high-recall candidate evidence set for each concern. The model then performs grounding and verification over this smaller candidate context and outputs the same schema as all other systems. Appendix~\ref{sec:agentic-details} provides the full pipeline details.

This pipeline improves GPT-4o-mini substantially over the single-shot setting: overall score increases from 0.266 to 0.403, coverage from 0.933 to 0.998, and manuscript grounding from 0.085 to 0.336. The improvement shows that explicit retrieval, decomposition, and schema normalization help models produce more complete and better-grounded predictions. 
% However, verification remains weak, increasing only to 0.300, indicating that workflow decomposition alone does not solve the core reasoning problem of deciding whether a revision adequately addresses a scientific concern. 
% The gain also comes at much higher inference cost: the agentic pipeline uses 91.9M tokens, compared with 20.0M tokens for single-shot GPT-4o-mini.

\subsection{Impact of Supervised Fine-tuning}
% 还需要说一下 sft 的训练数据
We further examine the effect of task-specific supervised fine-tuning by comparing Qwen3.5-9B with its supervised variant. 
500 papers are randomly sampled from the training dataset split for training.
Appendix~\ref{sec:sft-details} gives reproduce details, including the data construction, model, training, and inference details.

SFT Qwen3.5-9B improves overall performance from 0.463 to 0.614, with gains across all three components: characterization improves from 0.563 to 0.708, verification from 0.376 to 0.490, and grounding from 0.451 to 0.643. 
% 这里还需要提一下数据集的高质量，存在 knowledge，使得 sft 生效。
The improvements indicate that supervised adaptation helps models better capture the knowledge of \dataset{}, particularly the alignment between reviewer concerns, author responses, and manuscript evidence. The substantial gain in grounding further suggests that task-specific supervision improves evidence localization, enabling models to better identify relevant manuscript support for reviewer feedback.

\subsection{Ablation Study}
Table~\ref{tab:ablation} analyzes how different sources of information contribute to \dataset{} performance. We progressively remove components from the full-context setting to understand which capabilities depend on reviewer information, author responses, and manuscript evidence.

When only reviewer concerns are provided, GPT-5.5 achieves a relatively strong characterization score (0.676), but verification performance drops substantially (0.042), and grounding is not valid by design because no manuscript evidence is available. This indicates that reviewer text contains sufficient information to identify the type and scope of a concern, but is insufficient for determining whether the concern has been addressed.

Adding author responses improves verification to 0.236 while maintaining strong characterization (0.696). However, grounding remains unavailable without manuscript evidence. This suggests that response letters provide useful information about the claimed resolution, but these claims cannot be validated without examining the actual revision.

Providing concerns and manuscript inference substantially improves grounding, achieving 0.633 grounding and 0.582 overall performance. This demonstrates that manuscript evidence is the primary source for locating revision outcomes. Nevertheless, manuscript-only inference remains below full-context performance, particularly on verification (0.394 vs. 0.455), because the model lacks the original reviewer intent and the author's explanation of the change.

% Full-context inference achieves the strongest performance. These results show that reliable revision assessment requires jointly modeling the review concern, the author's response, and the revised manuscript. In particular, verification depends on connecting reviewer intent with concrete manuscript evidence rather than relying on any single information source.

\subsection{Closed-Loop Revision Case Study}
\label{sec:revision-loop-case-study}
We also test whether the benchmark is useful for choosing a verifier that improves a downstream revision loop. This case study asks whether systems that are stronger under the benchmark-style supervisor task provide more useful feedback when inserted between two author-revision calls. The pipeline starts from a initial submission draft, uses fixed human reviewer concerns from the review record, asks GPT-5.5 to produce a first revision, applies a ReviseBench-style supervisor to the first revision, renders the supervisor output as author-facing feedback, asks GPT-5.5 for a second revision, and finally uses a blinded Claude Opus 4.8 judge to compare concern-level outcomes.
We evaluate 10 sampled papers. The no-supervision condition receives the raw reviews and author-response context but no structured concern-level supervisor output.

\begin{table}[t]
\centering
% \scriptsize
\setlength{\tabcolsep}{4pt}
\begin{tabular}{lrrrr}
\toprule
Setting & Improved & Same & Worse & Net \\
\midrule
No supervision & 18.0\% & 66.5\% & 15.5\% & 0.025 \\
DeepSeek-V4-Flash & 29.2\% & 70.8\% & 0.0\% & 0.292 \\
GLM-5.2  & 35.4\% & 64.6\% & 0.0\% & 0.354 \\
\bottomrule
\end{tabular}
\caption{Closed-loop revision case study. Outcomes are concern-level pairwise comparisons between the first and second revision. Net is the improved minus the worse.}
\label{tab:revision-loop-case-study}
\end{table}

GLM-5.2 produces the largest number of useful second-revision changes. DeepSeek V4 Flash is weaker but still improves over the no-supervisor condition. Without supervisor feedback, GPT-5.5 still improves some concerns, but it also introduces substantially more regressions or less auditable changes. This supports the practical claim that \dataset{}-style verification is not only an offline scoring problem: better concern-resolution supervisors can provide more useful revision guidance in a closed authoring loop.

\section{Limitations}

% \paragraph{Granularity}
% \dataset{} evaluates revision outcomes at the level of structured concerns and manuscript evidence blocks. This formulation provides a reproducible evaluation target, but scientific revisions may involve complex changes spanning multiple sections or requiring broader contextual understanding. Future versions could explore richer evidence representations, including hierarchical or multi-span evidence annotations.

\paragraph{Data Diversity}
The current benchmark is based on Natural Communication transparent peer-review records, which mainly include papers that successfully completed the revision process. This may limit the diversity of revision trajectories and overrepresent well-addressed concerns. Future extensions could include broader venues and review outcomes to evaluate models under more diverse scenarios.

\paragraph{Task Boundary}
\dataset{} focuses on whether models can assess revision resolution from available evidence. In practice, additional capabilities such as generating revision suggestions, prioritizing reviewer concerns, and interacting with authors or reviewers are also important directions beyond the current benchmark scope.

% \paragraph{Annotation Scope}
% AutoSupervision evaluates whether reviewer concerns are connected to identifiable manuscript evidence after revision. It does not attempt to determine whether a revision is scientifically correct or whether a reviewer should ultimately accept a manuscript, as these judgments often require domain-specific expertise beyond textual evidence. Instead, the benchmark studies a narrower but important capability: evidence-grounded verification of revision claims.

% \paragraph{Evaluation Setting}
% \dataset{} focuses on evidence-based verification of manuscript revisions in transparent peer-review settings. This design enables precise evaluation with publicly available review artifacts, but it does not capture all aspects of real editorial workflows, such as confidential editor discussions or reviewer deliberations. Future extensions could incorporate additional sources of feedback to better model the broader publication process.

\section{Conclusion}
We introduced \dataset{}, a benchmark for scientific manuscript revision verification and concern-to-revision grounding. \dataset{} provides structured reviewer concerns, author responses, revised manuscript evidence, resolution labels, and exact evidence block annotations. 
Our evaluation shows that existing LLMs achieve strong characterization ability, but still struggle with verifying whether a revision genuinely addresses a reviewer concern and identifying the supporting manuscript evidence.

\bibliography{iclr2027_conference}
\bibliographystyle{iclr2027_conference}

\appendix
\newpage
\section{Prediction Schema and Evaluation Details}
\label{sec:prediction-evaluation-details}

Given a revision episode \(e_t=(M_t,R_t)\), the model produces one structured prediction for each review-response record \(r_{t,i}\). Each prediction consists of three groups of fields: \emph{Characterization}, \emph{Verification}, and \emph{Grounding}. Table~\ref{tab:prediction-schema} summarizes all scored fields and their valid options.

\paragraph{Characterization.}
Characterization evaluates whether a model correctly identifies the nature and scope of a reviewer comment. It includes three aspects: \texttt{comment\_kind}, which describes whether a comment is a critique, question, or suggestion; \texttt{point\_type}, which identifies the underlying issue category; and the target scope of the comment. We define \emph{target scope} as the joint specification of \texttt{target\_domain}, \texttt{target\_section}, and \texttt{target\_object}. Accordingly, \(S_{\mathrm{scope}}\) denotes the hierarchical target-scope score computed from these three fields.

\paragraph{Verification.}
Verification evaluates whether a model can determine whether a reviewer concern has been addressed in the revised manuscript. It consists of \texttt{paper\_status}, which describes the observed degree of change in the revised manuscript; \texttt{resolution\_label}, which indicates whether the concern is resolved, partially resolved, or unresolved; and \texttt{resolution\_evidence\_block\_ids}, which identifies the evidence blocks supporting the resolution judgment.

\paragraph{Grounding.}
Grounding evaluates whether a model can identify the exact revised-manuscript evidence supporting where a reviewer concern is addressed. It is measured using \texttt{target\_paper\_block\_ids}, which specifies the set of revised-manuscript evidence blocks where the concern is addressed or grounded in the paper.

\paragraph{Component scores.}
Categorical fields are evaluated using macro-F1 to reduce the influence of label-frequency imbalance. For ontology labels, macro-F1 is computed over labels with positive gold support. Concerns with the \texttt{unverifiable} resolution label are excluded according to the benchmark evaluation protocol.

Predicted evidence block ID sets are evaluated using micro-F1 because block identifiers are document-specific references rather than shared categorical labels.

The target scope is evaluated using \(S_{\mathrm{scope}}\), a hierarchical score that assigns partial credit according to the granularity of the matched target:

\[
S_{\mathrm{scope}}
=
\frac{1}{N}
\sum_{i=1}^{N}
\left(
0.2I_d+
0.4I_s+
0.4I_o
\right),
\]

where \(I_d\), \(I_s\), and \(I_o\) indicate exact matches of the target domain, target section, and target object, respectively.

The three component scores are:

\[
C_{\mathrm{char}}
=
\frac{1}{3}
\left(
F_{\mathrm{kind}}
+
F_{\mathrm{type}}
+
S_{\mathrm{scope}}
\right),
\]

\[
C_{\mathrm{ver}}
=
\frac{1}{3}
\left(
F_{\mathrm{res}}
+
F_{\mathrm{status}}
+
F_{\mathrm{evidence}}
\right),
\]

\[
C_{\mathrm{ground}}
=
F_{\mathrm{ground}}.
\]

Here, \(F_{\mathrm{kind}}\) and \(F_{\mathrm{type}}\) denote the macro-F1 scores for \texttt{comment\_kind} and \texttt{point\_type}, respectively. \(F_{\mathrm{res}}\) and \(F_{\mathrm{status}}\) denote the macro-F1 scores for \texttt{resolution\_label} and \texttt{paper\_status}. \(F_{\mathrm{evidence}}\) denotes the micro-F1 score for \texttt{resolution\_evidence\_block\_ids}, measuring whether the model identifies evidence blocks that support its resolution judgment. Finally, \(F_{\mathrm{ground}}\) denotes the micro-F1 score for \texttt{target\_paper\_block\_ids}, measuring whether the model identifies the exact revised-manuscript evidence blocks where the concern is addressed.

The overall benchmark score is the average of the three component scores:

\[
C_{\mathrm{overall}}
=
\frac{1}{3}
\left(
C_{\mathrm{char}}
+
C_{\mathrm{ver}}
+
C_{\mathrm{ground}}
\right).
\]

\begin{table*}[t]
\centering
\setlength{\tabcolsep}{4pt}
\begin{tabularx}{\textwidth}{@{}>{\raggedright\arraybackslash}l>{\raggedright\arraybackslash}X@{}}
\toprule
\textbf{Field} & \textbf{Options} \\
\midrule
\multicolumn{2}{@{}l}{\textbf{Characterization}} \\
Comment kind & critique; question; suggestion \\
Point type & novelty/significance; claim-evidence support; methodological soundness; experimental adequacy; baseline coverage; ablation coverage; statistical/reporting quality; reproducibility/implementation detail; writing/organization; figure/table presentation; theoretical/mechanistic justification; artifact/confound; clinical/application relevance; dataset/evaluation protocol; citation/related work; notation/internal consistency; supplementary/raw-evidence sufficiency; editorial/journal compliance; conclusion/generalization scope; limitation/scope \\
Target domain & main text; supplement \\
Target section & title; abstract; introduction; methods; results; discussion; conclusion; whole paper \\
Target object & text; figure; table; equation; reference; data \\
\midrule
\multicolumn{2}{@{}l}{\textbf{Verification}} \\
Paper status & clear change; partial change; no visible change \\
Resolution label & resolved; partially resolved; unresolved \\
Resolution evidence block IDs & Set of evidence block IDs supporting the resolution judgment. \\
\midrule
\multicolumn{2}{@{}l}{\textbf{Grounding}} \\
Target paper block IDs & Set of exact revised-manuscript evidence blocks where the concern is addressed or grounded in the paper. \\
\bottomrule
\end{tabularx}
\caption{Scored prediction fields and options. Target scope refers jointly to target domain, target section, and target object. Manuscript grounding refers to exact revised-manuscript evidence blocks supporting where a concern is addressed.}
\label{tab:prediction-schema}
\end{table*}

\section{Details of Dataset}

\subsection{Concern Attribute Distributions}
\label{sec:distribution}

Tables~\ref{tab:gold-distribution} and~\ref{tab:appendix-point-type-distribution}
summarize the concern-level composition of the final evaluation subset. As expected for accepted papers, most concerns are ultimately resolved, but the dataset still retains partially resolved and unresolved cases that require models to distinguish complete revision from incomplete response. The target-object distribution shows that revision verification is not a prose-only task: 1,185 concerns target figures, and additional concerns target data or analysis blocks, references, tables, and equations. The point-type distribution further shows that the benchmark spans a broad taxonomy of scientific revision needs, including writing and organization, claim-evidence support, statistical reporting, mechanistic justification, reproducibility details, figure/table presentation, experimental adequacy, methodology, scope, related work, and dataset or evaluation protocol concerns.

\begin{table}[ht]
\centering
\caption{Gold-label distributions in the full benchmark.}
\label{tab:gold-distribution}
% \small
\begin{tabular}{llr}
\toprule
Field & Label & Count \\
\midrule
comment\_kind & critique & 4,419 \\
comment\_kind & suggestion & 2,843 \\
comment\_kind & question & 1,528 \\
\midrule
resolution\_label & resolved & 8,265 \\
resolution\_label & partially resolved & 434 \\
resolution\_label & unresolved & 91 \\
\midrule
target\_object & text & 6,759 \\
target\_object & figure & 1,231 \\
target\_object & data & 365 \\
target\_object & reference & 276 \\
target\_object & table & 78 \\
target\_object & equation & 81 \\
\bottomrule
\end{tabular}
\end{table}

\begin{table*}[ht]
\centering
\caption{Point-type distribution in the full benchmark.}
\label{tab:appendix-point-type-distribution}
% \scriptsize
\setlength{\tabcolsep}{5pt}
\begin{tabular}{lr@{\hspace{10pt}}lr}
\toprule
Point type & Count & Point type & Count \\
\midrule
writing/organization & 1,242 & notation/internal consistency & 433 \\
claim-evidence support & 944 & methodological soundness & 412 \\
theoretical/mechanistic justification & 796 & conclusion/generalization scope & 399 \\
statistical/reporting quality & 786 & citation/related work & 383 \\
reproducibility/implementation detail & 706 & dataset/evaluation protocol & 355 \\
figure/table presentation & 643 & limitation/scope & 316 \\
experimental adequacy & 581 & artifact/confound & 290 \\
clinical/application relevance & 212 & novelty/significance & 96 \\
baseline coverage & 68 & supplementary/raw-evidence sufficiency & 63 \\
editorial/journal compliance & 39 & ablation coverage & 26 \\
\bottomrule
\end{tabular}
\end{table*}

\subsection{Agreement Experiment}
\label{sec:agreement}
Because the gold construction uses LLM-assisted structured extraction, we add an agreement check on 100 randomly sampled papers from the benchmark test split. The concern set is fixed to the GPT-5.5 output before running the second model.
It covers 160 revision episodes and 1,600 fixed concern instances. Table~\ref{tab:teacher-agreement} summarizes agreement between GPT-5.5 and Claude Opus 4.8. 
% The two models agree strongly on the auditable parts of the benchmark construction: target section reaches 0.873 exact agreement and 0.796 Cohen's $\kappa$.

\begin{table}[t]
\centering
% \scriptsize
\setlength{\tabcolsep}{4pt}
\begin{tabular}{lrrr}
\toprule
\multicolumn{4}{l}{\textbf{Categorical labels}} \\
Field & Acc. & $\kappa$ & AC1 \\
\midrule
Comment kind & 0.776 & 0.649 & 0.716 \\
Point type & 0.562 & 0.525 & 0.540 \\
Target section & 0.873 & 0.796 & 0.860 \\
Resolution label & 0.848 & 0.261 & 0.837 \\
\midrule
\multicolumn{4}{l}{\textbf{Evidence block sets}} \\
Field & Precision & F1 & IoU \\
\midrule
Target manuscript blocks & 0.773 & 0.710 & 0.673 \\
Resolution evidence & 0.648 & 0.441 & 0.308 \\
\bottomrule
\end{tabular}
\caption{Agreement on 100 sampled benchmark papers. Claude Opus 4.8 re-annotates fixed GPT-5.5 concern ids using the same structured schema.}
\label{tab:teacher-agreement}
\end{table}

The two models agree strongly on the benchmark construction.
Target section reaches 0.873 exact agreement and 0.796 Cohen's $\kappa$, concern taxonomy reaches 0.603 exact agreement and 0.540 $\kappa$, and the resolution label reaches 0.848 exact agreement.
For target manuscript evidence, the teachers achieve 0.710 block micro-F1 and 0.673 mean IoU. Resolution evidence is harder because it can include several adjacent rebuttal, follow-up, and confirmation blocks in the review record; across all such blocks, agreement reaches 0.441 micro-F1 and 0.308 mean IoU.

\section{Agentic Baseline Details}
\label{sec:agentic-details}

The agentic baseline uses the same base model, GPT-4o-mini, for all LLM calls and produces the same JSON prediction schema used by the single-shot baselines.

\paragraph{Concern characterization.}
The first stage reads the fixed review-response records for an episode and predicts concern-level characterization fields, including comment kind, point type, target domain, target section, target object, and response status. It also generates short search phrases used by the downstream manuscript retriever.

\paragraph{Deterministic manuscript-block retrieval.}
For each concern, it builds a query from the reviewer concern, author response, current-round discussion, the preliminary characterization labels, and the generated search phrases. It then scores every revised-manuscript block using lexical overlap, object cues such as figure/table/equation/reference mentions, and section/object consistency cues. The top 12 ranked blocks are retained as seed candidates, and a one-block neighbor window is added around each selected block to preserve local context. This top-12 setting is a recall-oriented candidate budget: these blocks are not the final predicted evidence, but the candidate pool from which the grounding stage selects final manuscript evidence IDs.

\paragraph{Grounding.}
The grounding stage receives each concern, its preliminary characterization, and the retrieved candidate manuscript blocks. It selects the target manuscript block IDs, predicts paper-side change status, and records whether manuscript grounding is available.

\paragraph{Verification.}
The verification stage receives the concern, author response, preliminary characterization, preliminary grounding, and selected manuscript evidence. It predicts response status, paper status, resolution label, and review-record evidence block IDs. If the grounding stage selects no manuscript blocks, the verification stage receives the top retrieved candidates as fallback context.

\paragraph{Normalization and cost.}
Finally, deterministic normalization merges the characterization, grounding, and verification outputs into one prediction per concern and enforces the benchmark schema. Grounding and verification are run in chunks of four concerns. The agentic pipeline improves structured-output reliability but is substantially more expensive than single-shot GPT-4o-mini, using 91.9M total tokens compared with 20.0M tokens for the single-shot baseline.

\section{Supervised Fine-tuning Details}
\label{sec:sft-details}

\paragraph{Data preparation.} The supervised model is trained from Qwen3.5-9B using the same structured prediction format as the main benchmark. The training data for SFT are generated from the dataset training split. We used 809 training episodes from 500 papers with 9,211 concern instances for SFT.

\paragraph{Training.}
Training was conducted on two NVIDIA H100 GPUs using full-parameter fine-tuning. The configuration uses the qwen3-5 model type, DeepSpeed ZeRO-3, bfloat16 precision, a maximum sequence length of 32,768, AdamW optimization, a learning rate of ($1\times10^{-5}$), cosine learning-rate scheduling, a warmup ratio of 0.03, weight decay of 0.1, gradient checkpointing, a per-device batch size of 1, gradient accumulation of 2, and a random seed of 42.

\paragraph{Inference and evaluation.} The checkpoint is served with an OpenAI-compatible vLLM endpoint, tensor parallel size 8, and maximum model length 32,768. 

\section{Reproducibility Statement}

In addition to the details we provide in Section \ref{sec:benchmark-construction} and Appendix \ref{sec:sft-details}, we also publicly release the benchmark dataset and code needed to reproduce the reported results.
% we also publicly release the benchmark dataset, construction scripts, evaluation code, and model prediction files needed to reproduce the reported results, subject to the redistribution policies of the source peer-review records.

\section{Ethics Statement}

\dataset{} is built from publicly available transparent peer-review records and published articles. These records were released by the source journal as part of its transparent review process, and our benchmark uses them only to study concern-level revision verification and evidence grounding.

\section{LLM Usage Statement}
LLMs were used to improve the grammar, wording, and readability of the manuscript. We also accessed commercial LLMs through their APIs as part of the benchmark evaluation, with a total API cost of approximately USD~2,000. The authors are fully responsible for the project design, experiments, analysis, citations, and final scientific claims.
% \fi

\end{document}